# Generalized Concomitant Multi-Task Lasso for sparse multimodal regression

Mathurin Massias[*1], Olivier Fercoq[†1], Alexandre Gramfort[‡2], and Joseph Salmon[§2]

[1]INRIA, Université Paris Saclay, 1 Rue Honoré d'Estienne d'Orves, 91120 Palaiseau, France
[2]LTCI, Télécom ParisTech, Université Paris-Saclay, 46 rue Barrault, 75013, Paris, France,

October 18, 2017

**Abstract**

In high dimension, it is customary to consider Lasso-type estimators to enforce sparsity. For standard Lasso theory to hold, the regularization parameter should be proportional to the noise level, yet the latter is generally unknown in practice. A possible remedy is to consider estimators, such as the Concomitant/Scaled Lasso, which jointly optimize over the regression coefficients as well as over the noise level, making the choice of the regularization independent of the noise level. However, when data from different sources are pooled to increase sample size, or when dealing with multimodal datasets, noise levels typically differ and new dedicated estimators are needed. In this work we provide new statistical and computational solutions to deal with such heteroscedastic regression models, with an emphasis on functional brain imaging with combined magneto- and electroencephalographic (M/EEG) signals. Adopting the formulation of Concomitant Lasso-type estimators, we propose a jointly convex formulation to estimate both the regression coefficients and the (square root of the) noise covariance. When our framework is instantiated to de-correlated noise, it leads to an efficient algorithm whose computational cost is not higher than for the Lasso and Concomitant Lasso, while addressing more complex noise structures. Numerical experiments demonstrate that our estimator yields improved prediction and support identification while correctly estimating the noise (square root) covariance. Results on multimodal neuroimaging problems with M/EEG data are also reported.

## 1 Introduction

In the context of regression, when the number of predictors largely exceeds the number of observations, sparse estimators provide interpretable and memory efficient models. Following the seminal work on the Lasso/Basis pursuit [Tibshirani, 1996, Chen and Donoho, 1995], a popular route to sparsity is to use convex $\ell_1$-type penalties. Lasso-type estimators rely on a regularization parameter $\lambda$ trading data-fitting versus sparsity, which requires careful tuning. Statistical analysis of the Lasso estimator states that $\lambda$ should be proportional to the noise level[1] [Bickel et al., 2009], though the latter is rarely known in practice. To address this issue, it has been proposed to jointly estimate the noise level along with the regression coefficients. A notable approach is via joint penalized maximum likelihood after a change of variable to avoid minimization of a non-convex function [Städler et al., 2010]. Another approach, the Concomitant

---

[*]`first.last@inria.fr`
[†]`first.last@telecom-paristech.fr`
[‡]`first.last@inria.fr`
[§]`first.last@telecom-paristech.fr`
[1]with Gaussian noise, the noise level stands for the standard deviation

Lasso [Owen, 2007] (inspired by Huber [1981]), and equivalent to the Square-root/Scaled Lasso [Belloni et al., 2011, Sun and Zhang, 2012]) includes noise level estimation by modifying the Lasso objective function. This estimator, which reaches optimal statistical rates for sparse regression [Belloni et al., 2011, Sun and Zhang, 2012], has the benefit that it makes the regularization parameter independent of the noise level. From a practical point of view, it is also well-suited for high dimension settings [Reid et al., 2016], and current solvers [Ndiaye et al., 2017] make its computation as fast as for the Lasso. While first attempts used second order cone programming solvers [Belloni et al., 2011], *e.g.,* TFOCS [Becker et al., 2011], recent ones rely on coordinate descent algorithms [Friedman et al., 2007] and safe screening rules [El Ghaoui et al., 2012, Fercoq et al., 2015].

In various applied contexts it is customary to pool observations from different sources or different devices, to increase sample size and boost statistical power. Yet, this leads to datasets with heteroscedastic noise. Heteroscedasticity, to be opposed to homoscedasticity, is a common statistical phenomenon occurring when observations are contaminated with non-uniform noise levels [Engle, 1982, Carroll and Ruppert, 1988]. This is for example the case of magneto- and electroencephalography (M/EEG) data, usually recorded from three types of sensors (gradiometers, magnetometers and electrodes), each having different signal and noise amplitudes.

Several statistical contributions have tried to address heteroscedastic models in high dimensional regression. Most works have relied on an exponential representation of the variance (the log-variance being modeled as a linear combination of the features), leading to non-convex objective functions. Solvers considered for such approaches require alternate minimization [Kolar and Sharpnack, 2012], possibly in an iterative fashion [Daye et al., 2012], a notable difference with a jointly convex formulation, for which one can control global optimality with duality gap certificates as proposed here. Similarly, Wagener and Dette [2012] estimate the variance thanks to a preliminary adaptive Lasso step, and use this information to correct the data-fitting term in a second step.

Here, we propose the multi-task Smoothed Generalized Concomitant Lasso (SGCL), an estimator that can handle data from different origins in a high dimensional sparse regression model by jointly estimating the regression coefficients and the noise levels of each modality or each data source. Contrary to other heteroscedastic Lasso estimators such as ScHeDs (a second order cone program) [Dalalyan et al., 2013], its computational cost is comparable to the Lasso, as it can benefit from coordinate descent solvers [Friedman et al., 2007] and other standard speed-ups popularized for the Lasso (safe rules [El Ghaoui et al., 2012], strong rules [Tibshirani et al., 2012], etc.). Besides, while this model provides sparse solutions, it also leads to a parameterization of the problem with one single scalar $\lambda$, that is independent of the multiple noise levels present in heterogeneous data.

Our manuscript is organized as follows. In Section 2, after reminding the necessary background of the Concomitant Lasso estimator, we introduce our general framework. We derive in Section 2 the necessary mathematical results to obtain an efficient solver based on coordinate descent. We then present in Section 4 a lightweight version of it, adapted to more specific noise models. Finally, in Section 5 we provide empirical evidence using simulations with known ground truth that our model yields better support recovery and prediction than homoscedastic estimators. On the problem of source localization with real M/EEG recordings, we show that the proposed model leads to consistent estimators of the noise standard deviations for each modality, hence learning from the data the right balance between complementary modalities with high or low SNRs.

## 2 Concomitant estimators

**Notation** For any integer $d \in \mathbb{N}$, we denote by $[d]$ the set $\{1,\ldots,d\}$. Our observation matrix is $Y \in \mathbb{R}^{n \times q}$ with $n$ the number of samples, with $q$ the number of tasks and the design matrix $X = [X_1,\ldots,X_p] \in \mathbb{R}^{n \times p}$ has $p$ explanatory variables or features, stored column-wise. The standard Euclidean norm (resp. inner product) on vectors or matrices is written $\|\cdot\|$ (resp. $\langle \cdot, \cdot \rangle$), the $\ell_1$ norm $\|\cdot\|_1$, the $\ell_\infty$ norm $\|\cdot\|_\infty$, and the matrix transposition of a matrix $Q$ is denoted by $Q^\top$. For $B \in \mathbb{R}^{p \times q}$, its $j^{th}$ row is $B_{j,:}$ and $\|B\|_{2,1} = \sum_{j=1}^{p} \|B_{j,:}\|$ (resp. $\|B\|_{2,\infty} = \max_{j \in [p]} \|B_{j,:}\|$) is its row-wise $\ell_{2,1}$ (resp. $\ell_{2,\infty}$) norm. For matrices, $\|\cdot\|_2$ is the spectral norm. For real numbers $a$ and $b$, $a \vee b$ stands for the maximum of $a$ and $b$, and $(a)_+ = a \vee 0$.

We denote $\mathrm{BST}(\cdot, \tau)$ the block soft-thresholding operator at level $\tau > 0$, i.e., $\mathrm{BST}(x, \tau) = (1 - \tau/\|x\|)_+ \cdot x$ for any $x \in \mathbb{R}^d$ (with the convention $\frac{0}{0} = 1$).

The sub-gradient of a convex function $f : \mathbb{R}^d \to \mathbb{R}$ at $x$ is defined as $\partial f(x) = \{z \in \mathbb{R}^d : \forall y \in \mathbb{R}^d, f(y) - f(x) \geq \langle z, y - x \rangle\}$. We denote by $\iota_C$ the indicator function of a set $C$, defined as $\iota_C(x) = 0$ if $x \in C$ and $\iota_C(x) = +\infty$ if $x \notin C$. The identity matrix of size $n \times n$ is denoted by $\mathrm{I}_n$ (or I, when there is no dimension ambiguity).

We write $\mathbb{S}^n$ for the set of symmetric matrices and $\mathbb{S}^n_+$ (resp. $\mathbb{S}^n_{++}$) for the set of positive semi-definite matrices (resp. positive definite matrices). For two matrices $S_1$ and $S_2$ we write $S_1 \succeq S_2$ (resp. $S_1 \succ S_2$) for $S_1 - S_2 \in \mathbb{S}^n_+$ (resp. $S_1 - S_2 \in \mathbb{S}^n_{++}$). The symbol Tr denotes the trace operator, and $\|A\|_S = \sqrt{\mathrm{Tr}\, A^\top S A}$ is the Mahalanobis norm induced by $S \in \mathbb{S}^n_{++}$. For more compact notation, for $\underline{\sigma} > 0$ we denote $\underline{\Sigma} = \underline{\sigma}\, \mathrm{I}_n$.

As much as possible, we denote vectors with lower case letters and matrices with upper case ones.

## 2.1 Reminder on Concomitant Lasso

Let us first recall the Concomitant Lasso estimator, following the vector formulation ($y \in \mathbb{R}^n$) proposed in [Owen, 2007, Sun and Zhang, 2012].

**Definition 1.** For $\lambda > 0$, the Concomitant Lasso coefficient and standard deviation estimators are defined as solutions of the optimization problem

$$\arg\min_{\beta \in \mathbb{R}^p, \sigma > 0} \frac{\|y - X\beta\|^2}{2n\sigma} + \frac{\sigma}{2} + \lambda \|\beta\|_1 \,. \tag{2.1}$$

To avoid numerical issues when $\sigma$ approaches 0, it was proposed in Ndiaye et al. [2017] to add a constraint on $\sigma$ in the objective function. Following the terminology introduced in Nesterov [2005], this was coined the Smoothed Concomitant Lasso.

**Definition 2.** For $\underline{\sigma} > 0$ and $\lambda > 0$, the Smoothed Concomitant Lasso estimator $\hat{\beta}$ and its associated standard deviation estimator $\hat{\sigma}$ are defined as

$$(\hat{\beta}, \hat{\sigma}) \in \arg\min_{\beta \in \mathbb{R}^p, \sigma \geq \underline{\sigma}} \frac{\|y - X\beta\|^2}{2n\sigma} + \frac{\sigma}{2} + \lambda \|\beta\|_1 \,. \tag{2.2}$$

## 2.2 General problem formulation

Motivated by our application to the M/EEG inverse problem, we present all the results in a multi-task setting. These results are still valid for single task problems ($q = 1$), in which case the formulas and algorithms are often simpler. A special focus on the case $q = 1$ is provided in Appendix B.

We now extend the Smoothed Concomitant Lasso to more general noise models, and present some properties obtained thanks to convexity and duality. As a **warning**, in our formulation the matrix $\Sigma^* \in \mathbb{S}^n_{++}$ is the co-standard-deviation matrix (the square-root of the covariance matrix), in contrast with the standard Gaussian noise model notation. The model reads:

$$Y = XB^* + \Sigma^* E \,, \tag{2.3}$$

where entries of E are independent, centered and normally distributed.

**Definition 3.** For $\underline{\sigma} > 0$ and $\lambda > 0$ (recall that we denote $\underline{\Sigma} = \underline{\sigma} \, I_n$), we define the multi-task Smoothed Generalized Concomitant Lasso (multi-task SGCL) estimator $\hat{B}$ and its associated co-standard-deviation matrix $\hat{\Sigma}$ as the solutions of the optimization problem

$$(\hat{B}, \hat{\Sigma}) \in \underset{B \in \mathbb{R}^{p \times q}, \Sigma \in \mathbb{S}_{++}^n, \Sigma \succeq \underline{\Sigma}}{\arg\min} \mathcal{P}^{(\lambda)}(B, \Sigma) \enspace, \tag{2.4}$$

with $\mathcal{P}^{(\lambda)}(B, \Sigma) = \dfrac{\|Y - XB\|_{\Sigma^{-1}}^2}{2nq} + \dfrac{\text{Tr}(\Sigma)}{2n} + \lambda \|B\|_{2,1}$.

**Remark 1.** Concomitant estimators such as the Smoothed Concomitant Lasso rely on perspective functions. A general framework for optimization with similar functions is provided in Combettes and Müller [2016], and applies for instance to other (potentially non-convex) alternative noise estimators, *e.g.*, TREX [Lederer and Müller, 2015]. Yet, we are not aware of a matrix perspective theory as we propose in the present work to handle anisotropic noise.

**Proposition 1.** *The formulation given in* (2.4) *is jointly convex.*

*Proof.* The constraint set is convex and the matrix-fractional function $(Z, \Sigma) \mapsto \text{Tr}\, Z^\top \Sigma^{-1} Z$ is jointly convex over $\mathbb{R}^{n \times q} \times \mathbb{S}_{++}^n$, *cf.* Boyd and Vandenberghe [2004, Example 3.4]. □

As for $\underline{\sigma}$ in the Smoothed Concomitant Lasso, the constraint $\Sigma \succeq \underline{\Sigma}$ acts as a regularizer in the dual, and it is introduced for numerical stability. In practice, the value of $\underline{\Sigma} = \underline{\sigma} \, I_n$ can be set as follows:

- If prior information on the minimal noise level present in the data is available, $\underline{\sigma}$ can be set as this bound. Indeed, if $\hat{\Sigma} \succ \underline{\Sigma}$, then the constraint $\Sigma \succeq \underline{\Sigma}$ is not active and the solution to (2.4) is a solution of the non-smoothed problem with $\underline{\Sigma} = 0$.

- Without prior information on the noise level, one can use a proportion of the initial estimation of the noise standard deviation $\underline{\sigma} = 10^{-\alpha} \|Y\|/\sqrt{nq}$, with for example $\alpha \in \{2, 3\}$.

## 3 Properties of the multi-task SGCL

**Theorem 1.** *The dual formulation of the multi-task Smoothed Generalized Concomitant Lasso reads*

$$\hat{\Theta} = \underset{\Theta \in \Delta_{X,\lambda}}{\arg\max} \underbrace{\langle Y, \lambda \Theta \rangle + \underline{\sigma} \left( \dfrac{1}{2} - \dfrac{nq\lambda^2}{2} \|\Theta\|^2 \right)}_{\mathcal{D}^{(\lambda, \underline{\Sigma})}(\Theta)} \enspace, \tag{3.1}$$

*for* $\Delta_{X,\lambda} = \left\{ \Theta \in \mathbb{R}^n : \|X^\top \Theta\|_{2,\infty} \le 1, \|\Theta\|_2 \le \dfrac{1}{\lambda n \sqrt{q}} \right\}$.

*The link between $\hat{B}$ and $\hat{\Sigma}$ is detailed in Proposition 2. We also have the link-equation between primal and dual solutions:*

$$\hat{\Theta} = \dfrac{1}{nq\lambda} \hat{\Sigma}^{-1} (Y - X\hat{B}) \enspace, \tag{3.2}$$

*and the sub-differential inclusion:*

$$X^\top \hat{\Sigma}^{-1}(Y - X\hat{B}) \in nq\lambda \partial \|\cdot\|_{2,1}(\hat{B}) \enspace. \tag{3.3}$$

The proof of these results is in Appendix A.1.

**Remark 2.** The link equation provides a natural way to construct a dual feasible point from any pair $(B, \Sigma)$. Since at convergence Equation (3.2) holds, we can choose as a dual point $\Theta = \Sigma^{-1}(Y - XB)/\alpha$ where $\alpha = \|X^\top \Sigma^{-1}(Y - XB)\|_{2,\infty} \vee \lambda n \sqrt{q} \|\Sigma^{-1}(Y - XB)\|_2$ is a scalar chosen to make $\Theta$ dual feasible.

**Proposition 2.** *The solution of*

$$\Sigma \in \underset{\Sigma \in \mathbb{S}_{++}^n, \Sigma \geq \underline{\Sigma}}{\arg\min} \frac{1}{2nq} \|Y - XB\|_{\Sigma^{-1}}^2 + \frac{1}{2n} \operatorname{Tr}(\Sigma) \ , \tag{3.4}$$

*is given by*

$$\Sigma = \Psi(Z, \underline{\sigma}) := U \operatorname{diag}(\mu_1, \ldots, \mu_r, \underline{\sigma}, \ldots, \underline{\sigma}) U^\top \ , \tag{3.5}$$

*where* $Z = \frac{1}{\sqrt{q}}(Y - XB)$, $U \operatorname{diag}(\lambda_1, \ldots, \lambda_r, 0, \ldots, 0) U^\top$ *is an eigenvalue decomposition of* $ZZ^\top$, *(i.e., r is the rank of* $ZZ^\top$, $\lambda_1 \geq \cdots \geq \lambda_r > 0$ *and* $UU^\top = I_n$*), and for* $i \in [r]$, $\mu_i = \sqrt{\lambda_i} \vee \underline{\sigma}$.

The proof is included in Appendix A.2.

**Remark 3.** Formula (3.5) makes it straightforward to compute $\Sigma^{-1}$ and $\operatorname{Tr} \Sigma$, which we rather store than $\Sigma$ for computational efficiency in Algorithm 1.

At every update of $\Sigma$, it is also beneficial to precompute $\Sigma^{-1}X$ and $\Sigma^{-1}R$: maintaining $\Sigma^{-1}R$ rather than $R$ enables to avoid multiplication by $\Sigma^{-1}$ at every block-coordinate step.

**Remark 4.** Similarly to the Concomitant Lasso, and contrary to the Lasso, the multi-task SGCL is equivariant under scaling of the response, in the following sense: consider the transformation

$$Y' = \alpha Y, \ B' = \alpha B, \ \Sigma' = \alpha \Sigma, \quad (\alpha > 0) \ ,$$

which leaves the model (2.3) invariant. Then one can check that the solutions of (2.4) are multiplied by the same factor: $\hat{B}' = \alpha \hat{B}$ and $\hat{\Sigma}' = \alpha \hat{\Sigma}$

## 3.1 Critical parameter

As for the Lasso, the null vector is optimal for the multi-task Smoothed Generalized Concomitant Lasso problem as soon as the regularization parameter becomes too large. We refer to the smallest $\lambda$ leading to a null solution as the *critical parameter* and denote it $\lambda_{\max}$. Its computation is detailed in the next proposition.

**Proposition 3.** *For the multi-task SGCL estimator we have the following property: with* $\Sigma_{\max} = \Psi(Y, \underline{\sigma})$,

$$\hat{B} = 0, \quad \forall \lambda \geq \lambda_{\max} := \frac{1}{nq} \|X^\top \Sigma_{\max}^{-1} Y\|_{2,\infty} \ . \tag{3.6}$$

*Proof.* Fermat's rule for (2.4) states that

$$(0, \hat{\Sigma}) \in \underset{B \in \mathbb{R}^{p \times q}, \Sigma \in \mathbb{S}_{++}^n, \Sigma \geq \underline{\Sigma}}{\arg\min} \mathcal{P}(B, \Sigma)$$

$$\Leftrightarrow 0 \in \left\{-\frac{1}{nq} X^\top \hat{\Sigma}^{-1} Y\right\} + \lambda \mathcal{B}_{2,\infty}$$

$$\Leftrightarrow \frac{1}{nq} \left\|X^\top \hat{\Sigma}^{-1} Y\right\|_{2,\infty} \leq \lambda.$$

Thus, $\lambda_{\max} = \frac{1}{nq} \|X^\top \hat{\Sigma}^{-1} Y\|_{2,\infty}$ is the critical parameter. Then, notice that for $\hat{B} = 0$ one has

$$\hat{\Sigma} = \Psi(Y, \underline{\sigma}) = \underset{\Sigma \in \mathbb{S}_{++}^n, \Sigma \geq \underline{\Sigma}}{\arg\min} \frac{1}{2nq} \|Y\|_{\Sigma^{-1}}^2 + \frac{1}{2n} \operatorname{Tr}(\Sigma) \ .$$

□

**Algorithm 1** ALTERNATE MIN. FOR MULTI-TASK SGCL
___
**input** : $X, Y, \underline{\Sigma}, \lambda, f, T$
**init** : $B = 0_{p,q}, \Sigma^{-1} = \underline{\Sigma}^{-1}, R = Y$
**for** iter $= 1, \ldots, T$ **do**
    **if** iter $= 1 \pmod{f}$ **then**
        $\Sigma \leftarrow \Psi(R, \underline{\Sigma})$                                                                    // update of Proposition 2
        **for** $j = 1, \ldots, p$ **do**
            $L_j = X_j^\top \Sigma^{-1} X_j$
    **for** $j = 1, \ldots, p$ **do**
        $R \leftarrow R + X_j B_j$                                                                        // partial residual update
        $B_j \leftarrow \mathrm{BST}\left(\dfrac{X_j^\top \Sigma^{-1} R}{L_j}, \dfrac{\lambda n q}{L_j}\right)$                                            // coef. update
        $R \leftarrow R - X_j B_j$                                                                       // residual update
**return** $B, \Sigma$
___

## 3.2 Algorithm

Since the multi-task SGCL formulation is jointly convex, one can rely on alternate minimization to find a solution. Moreover, the formulation has the appealing property that when $\Sigma$ is fixed, the convex problem in B is a standard "smooth + $\ell_1$-type" problem. This can be solved easily using standard block coordinate descent (BCD) algorithm popularized for Lasso solvers by Friedman et al. [2007]. Alternatively, when B is fixed, the minimization in $\Sigma$ has the closed-form solution of Proposition 2. The $\Sigma$ update being more costly than the B update, one can perform it every $f$ BCD epochs (*i.e.*, if $f = 10$, every ten passes over the $p$ rows $B_{j,.}$). This minimization scheme is summarized in Algorithm 1, and details of the updates formulas are given in Appendix A.3.

Since strong duality holds for (2.4), we use the duality gap as a stopping criterion for the convergence. Every $f$ epochs of coordinate descent as presented in Algorithm 1, we compute a dual point $\Theta$ (see Remark 2), evaluate the duality gap, and stop if it is lower than $\epsilon = 10^{-6}/\|Y\|$. The pair $(B, \Sigma)$ obtained when the duality gap goes below $\epsilon$ is then guaranteed to be an $\epsilon$-solution of (2.4).

## 3.3 Statistical limits

We provided the most general framework to adapt the Concomitant Lasso to multi-task and non scalar covariances.

However, in its general formulation the multi-task Smoothed Generalized Concomitant Lasso has an obvious drawback: in practice estimating $\Sigma^*$ requires to fit $n(n-1)/2$ parameters with only $nq$ observations, which is problematic if $q$ is not large enough. Hence, additional regularization might be needed to provide an accurate estimator of $\Sigma^*$, *e.g.*, following the direction proposed in Ledoit and Wolf [2004].

Nevertheless, in common practical scenarios $\Sigma^*$ can be assumed to have a more regular structure. In the following, we address in details a special case of heteroscedastic models, where the noise is de-correlated and has a known block-wise structure.

# 4 Block homoscedastic model

Motivated by the specificities of the M/EEG inverse problem (see Section 5.3), but more generally by supervised learning problems where data come from an identified, finite set of sources, we now propose a specification of (2.3) when more assumptions can be made about the noise. Indeed, when the observations come from $K$ different sources or $K$ types of sensors (in the M/EEG case: magnetometers, gradiometers and electrodes), $\Sigma$ can be estimated in a simplified way. Assuming independent noise among data sources or sensor types, we propose a variant of (2.3), called the *block homoscedastic* model. In this model, $\Sigma^*$ is constrained to be diagonal, the diagonal being constant over known blocks.

Formally, if the $k$-th group of sensors is composed of $n_k$ sensors ($\sum_1^K n_k = n$), with design matrix $X^k \in \mathbb{R}^{n_k \times p}$, observation matrix $Y^k \in \mathbb{R}^{n_k \times q}$ and noise level $\sigma_k^* > 0$, the block homoscedastic model is a combination of $K$ homoscedastic models: $\forall k \in [K], \quad Y^k = X^k B^* + \sigma_k^* E^k$, with the entries of $E^k$ independently sampled from $\mathcal{N}(0,1)$. In the following we denote

$$X = \begin{pmatrix} X^1 \\ \vdots \\ X^K \end{pmatrix}, Y = \begin{pmatrix} Y^1 \\ \vdots \\ Y^K \end{pmatrix}, E = \begin{pmatrix} E^1 \\ \vdots \\ E^K \end{pmatrix}, \text{ and } \Sigma^* = \begin{pmatrix} \sigma_1^* I_{n_1} & & 0 \\ & \ddots & \\ 0 & & \sigma_K^* I_{n_K} \end{pmatrix} \in \mathbb{S}_{++}^n .$$

Following this model, we call multi-task Smoothed Block Homoscedastic Concomitant Lasso (multi-task SBHCL) the estimator similar to (2.4) with the additional constraint that $\Sigma$ is a diagonal matrix $\mathrm{diag}(\sigma_1 I_{n_1}, \ldots, \sigma_K I_{n_K})$, with $K$ constraints $\sigma_k \geq \underline{\sigma}_k$:

$$\underset{\substack{B \in \mathbb{R}^{p \times q}, \\ \sigma_1, \ldots, \sigma_K \in \mathbb{R}_{++}^K \\ \sigma_k \geq \underline{\sigma}_k, \forall k \in [K]}}{\arg\min} \sum_{k=1}^K \left( \frac{\|Y^k - X^k B\|^2}{2nq\sigma_k} + \frac{n_k \sigma_k}{2n} \right) + \lambda \|B\|_{2,1} . \quad (4.1)$$

Since no closed-form solution is available for (4.1), we also propose an iterative solver (see Section 3.2), and we again propose a stopping condition based on the duality gap, which is derived for this problem in Appendix A.4.

- When the constraints on the $\sigma_k$'s are not saturated at optimality, formulation (4.1) has an equivalent square-root Lasso [Belloni et al., 2011] formulation: $\arg\min_{B \in \mathbb{R}^{p \times q}} \frac{1}{nq} \sum_{k=1}^K \sqrt{n_k} \|Y^k - X^k B\| + \lambda \|B\|_{2,1}$.
- To fix the values of the lower bounds on the noise levels $\sigma_k$, we use an arbitrary proportion of the initial estimation of the noise variances per block *i.e.*, $\underline{\Sigma} = 10^{-\alpha} \mathrm{diag}(\|Y^1\|/\sqrt{n_1 q} I_{n_1}, \ldots, \|Y^K\|/\sqrt{n_K q} I_{n_K})$. $\alpha = 3$ is used in the experiments.

The equivalents of Theorem 1, Proposition 2 and Proposition 3 for the multi-task SBHCL are:

**Theorem 2.** *The dual formulation of* (4.1) *is*

$$\hat{\Theta} = \underset{\Theta \in \Delta'_{X,\lambda}}{\arg\max} \langle Y, \lambda \Theta \rangle + \sum_{k=1}^K \frac{\underline{\sigma}_k}{2} \left( \frac{n_k}{n} - nq\lambda^2 \|\Theta^k\|^2 \right) ,$$

*where* $\Delta'_{X,\lambda}$ *is defined by*

$$\Delta'_{X,\lambda} = \Big\{ \Theta \in \mathbb{R}^{n \times q} :$$
$$\|X^\top \Theta\|_{2,\infty} \leq 1, \forall k \in [K], \|\Theta^k\| \leq \frac{\sqrt{n_k}}{n\lambda\sqrt{q}} \Big\}.$$

**Proposition 4.** *When optimizing* (4.1) *with* $\hat{B}$ *being fixed, then* $\hat{\Sigma} = \mathrm{diag}(\hat{\sigma}_1 I_{n_1}, \ldots, \hat{\sigma}_K I_{n_K})$, *with residuals* $R^k = Y^k - X^k \hat{B}$ *and* $\hat{\sigma}_k = \underline{\sigma}_k \vee (\|R^k\|/\sqrt{n_k q})$.

**Proposition 5.** *For the multi-task SBHCL the critical parameter is* $\lambda_{\max} := \|X^\top \Sigma_{\max}^{-1} Y\|_{2,\infty}/n$ *where* $\Sigma_{\max} = \mathrm{diag}(\sigma_1^{\max} I_{n_1}, \ldots, \sigma_K^{\max} I_{n_K})$ *and* $\forall k \in [K], \sigma_k^{\max} = \underline{\sigma}_k \vee (\|Y^k\|/\sqrt{n_k q})$.

The proofs are similar to those of Proposition 2 and Proposition 3 and delayed to Appendix A.4.

The strategy of Algorithm 1 can also be applied to the multi-task SBHCL. Because of the special form of $\Sigma$, the computations are lighter and the standard deviations $\sigma_k$'s can be updated at each coordinate

**Algorithm 2** ALTERNATE MIN. FOR MULTI-TASK SBHCL

**input** : $X^1, \ldots, X^K, Y^1, \ldots, Y^K, \underline{\sigma}_1, \ldots, \underline{\sigma}_K, \lambda, T$
**init** : $B = 0_{p,q}, \forall k \in [K], \sigma_k = \|Y^k\|/\sqrt{n_k q}, R^k = Y^k, \forall k \in [K], \forall j \in [p], L_{k,j} = \|X_j^k\|_2^2$

**for** iter $= 1, \ldots, T$ **do**
    **for** $j = 1, \ldots, p$ **do**
        **for** $k = 1, \ldots, K$ **do**
            $R^k \leftarrow R^k + X_j^k B_j$                                                                        // residual update
        $B_j \leftarrow \mathrm{BST}\left(\sum_{k=1}^{K} \frac{X_j^{k\top} R^k}{\sigma_k}, \lambda nq\right) / \sum_{k=1}^{K} \frac{L_{k,j}}{\sigma_k}$                          // soft-thresholding
        **for** $k = 1, \ldots, K$ **do**
            $R^k \leftarrow R^k - X_j^k B_j$                                                                        // residual update
            $\sigma_k \leftarrow \underline{\sigma}_k \vee \frac{\|R^k\|}{\sqrt{n_k q}}$                                                          // std dev update
**return** $B, \sigma_1, \ldots, \sigma_k$

descent update. Indeed, updating all the $\sigma_k$'s may seem costly, since a naive implementation requires to recompute all the residual norms $\|R^k\|$, where $R^k = Y^k - X^k B$, which is $\mathcal{O}(nq)$. However, it is possible to store the values of $\|R^k\|^2$ and update them at each $B_j$ update with a $\mathcal{O}(kq)$ cost. Indeed, if we denote $\tilde{B}_j$ and $\tilde{R}^k$ the values before the update, we have:

$$R^k = \tilde{R}^k + X_j^{k\top}(\tilde{B}_j - B_j)$$
$$\left\|R^k\right\|^2 = \|\tilde{R}^k\|^2 + 2\,\mathrm{Tr}[(\tilde{B}_j - B_j)\tilde{R}^{k\top} X_j^k]$$
$$\qquad + \|\tilde{B}_j - B_j\|^2 L_{j,k}$$

and all the quantities $\tilde{R}^{k\top} X_j^k$ are already computed for the soft-thresholding step. As $k \leq n$, this makes the cost of one $B_j$ update of Algorithm 2 $\mathcal{O}(nq)$, the same cost as for the $\ell_{2,1}$ regularized Lasso, *a.k.a.* multi-task Lasso (MTL) [Obozinski et al., 2010].

## 5 Experiments

To demonstrate the benefits of handling non-homoscedastic noise, we now present experiments using both simulations and real M/EEG data. First, we show that taking into account multiple noise levels improves both prediction performance and support identification. We then illustrate on M/EEG data that the estimates of the noise standard deviations using multi-task SBHCL match the expected behavior when increasing the SNR of the data. We also demonstrate empirically the benefit of our proposed multi-task SBHCL to reduce the variance of the estimation.

We consider the case where the block structure of the noise is known by the practitioner. Therefore, all experiments use the block homoscedastic setting. Note that this is relevant with the M/EEG framework where the variability of the noise is due to different data acquisitions sensors that are known.

### 5.1 Prediction performance

We first study the impact of the multi-task SBHCL on prediction performance, evaluated on left-out data.

The experiment setup is as follows. There are $n = 300$ observations, $p = 1,000$ features and $q = 100$ tasks. The design $X$ is random with Toeplitz-correlated features with parameter $\rho = 0.7$ (correlation between features $i$ and $j$ is $\rho^{|i-j|}$). The true coefficient matrix $B^*$ has 20 non-zero rows, whose entries are independently and normally (centered and reduced) distributed. We simulate data coming from $K = 3$ sources (each one containing 100 observations) whose respective noise levels are $\sigma^\star$, $2\sigma^\star$ and $5\sigma^\star$. The standard deviation $\sigma^\star$ is chosen so that the signal-to-noise ratio

$$\text{SNR} := \|Y\|/\|XB^*\| = 1 \ .$$

The two estimators are trained for $\lambda$ varying on a logarithmic grid of 15 values between the critical parameter[2] $\lambda_{max}$ and $\lambda_{max}/10$. The training set contains 150 samples ($n_1 = n_2 = n_3 = 50$ of each data source) and the test set consists of the remaining 150.

Figure 1 shows prediction performance for the Smooth Concomitant Lasso (SCL), which estimates a single noise level for all blocks, and the multi-task SBHCL. Since each block has a different noise level, for each block $k$ and each estimator, we report the Root Mean Squared error (RMSE, $\|Y^k - X^k\hat{B}\|/\sqrt{qn_k}$) normalized by the oracle RMSE ($\|Y^k - X^kB^*\|/\sqrt{qn_k}$). After taking the log, zero value means a perfect estimation, a positive value means under-fitting of the block, while a negative value corresponds to over-fitting. Figure 1 reports normalized RMSE values on both the training and the test data.

As it can be observed, the RMSE for the multi-task SBHCL is lower on every block of the test set, meaning that it has better prediction performance. By attributing a higher noise standard deviation to the noisiest block (block 3), the multi-task SBHCL is able to down-weight the impact of these samples on the estimation, while still benefiting from it.

While the 3 normalized RMSE have similar behaviors on the test set for the SCL, for low values of $\lambda$, the multi-task SBHCL overfits more on the least noisy block. However this does not result in degraded prediction performance on the test set, neither for this block nor for others, and the prediction is even better on the noisiest block. Indeed, the SCL overfits more on the the noisiest block, which has a greater impact on prediction (as overfitting on noiseless data would lead to perfect parameter inference). When the regularization parameter becomes too low, taking into account different noise levels allows our estimator to limit the impact of overfitting by favoring the most reliable source. This experiments shows that our formulation is appealing for parameter selection, as the best left-out prediction is obtained for similar values of $\lambda$.

## 5.2 Support recovery

In this experiment, we demonstrate the superior performance of the multi-task SBHCL for support recovery, *i.e.*, its ability to correctly identify the predictive features. The experimental setup is the same as in Section 5.1, except that the support of $B^*$ is of size 50. We also vary $\rho \in \{0.1, 0.9\}$ and the SNR $\in \{1, 5\}$ (additional results are included in Appendix C).

The five estimators compared on Figure 2 are the multi-task SBHCL, the SCL, the MTL, and also the MTL and the SCL trained on the least noisy block (*i.e.*, the most favorable block). Following the empirical evaluation from [Bühlmann and Mandozzi, 2014], the figure of merit is the ROC curve, *i.e.*, the true positive rate as a function of the false positive rate. The curve is obtained by varying the value of the regularization parameter $\lambda$ (lower values leading to larger predicted support and therefore potentially more false positives).

We can see that when the SNR is sufficiently high (top graph with $SNR = 5$), the multi-task SBHCL, the SCL and the MTL successfully recover the true support, while the MTL or SCL trained on the least noisy block with only one third of the data fails. However, when the SNR is lower (middle graph with SNR = 1), the multi-task SBHCL still achieves almost perfect support identification, while the performance of the MTL and SCL decreases. The performance is naturally even worse when using only one block of samples. Finally, when the features are more correlated ($\rho = 0.9$) and the conditioning of $X$ is degraded, the multi-task SBHCL, despite not perfectly recovering the true support, still has superior performance. Note also that unsurprisingly the MTL and the SCL lead to almost perfectly the same ROC curves as both estimators (if $\underline{\sigma}$ is small enough) have the same solution path. Any difference between SCL and MTL in our graph is due to the choice of a discrete set of $\lambda$ values.

---

[2]Note that $\lambda_{max}$ is model specific

### 5.3 Results on joint M/EEG real data during auditory stimulation

We now evaluate our estimator on magneto- and electroencephalography (M/EEG) data.

The data consists of M/EEG recordings, which measure the electric potential and magnetic field induced by active neurons. Data are time-series so that $n$ corresponds to the number of sensors and $q$ to the number of consecutive time instants in the data. Thanks to their high temporal resolution on the order of milliseconds, M/EEG help to elucidate where and precisely when cognitive processes happen in the brain [Baillet, 2017]. The so-called M/EEG inverse problem, which consists in identifying active brain regions, can be cast as a high-dimensional sparse regression problem.

Because of the limited number of sensors, as well as the physics of the problem, the M/EEG inverse problem is severely ill-posed, and regularization is needed to provide solutions which are both biologically plausible and robust to measurement noise [Wipf et al., 2008, Haufe et al., 2008, Gramfort et al., 2013]. As foci of neural activity are observed from a distance by M/EEG and since only a small number of brain regions are involved in a cognitive task during a short time interval, it is common to employ regularizations that promote sparsity. Amongst these, the $\ell_1/\ell_2$ penalty has been successfully applied to the M/EEG inverse problem in either time [Ou et al., 2009] or frequency domain [Gramfort et al., 2013].

The experimental condition considered is a monaural auditory stimulation in the right ear of the subject. The same subject undergoes the same stimulation 61 times, and the M/EEG measurements are recorded from 0.2 s before to 0.5 s after the auditory stimulus. The dataset (from the MNE software [Gramfort et al., 2014]) thus contains 61 repetitions (*trials*) of this stimulation.

In the experimental setup we have 204 gradiometers, 102 magnetometers and 60 EEG electrodes. We have discarded one magnetometer and one electrode corrupted by strong artifacts. We therefore have $K = 3$ sensor types with $n_1 = 203$, $n_2 = 102$ and $n_3 = 59$ (so $n = 364$). The design matrix $X$ is obtained by numerically solving the M/EEG forward problem using $p = 1884$ candidate sources regularly distributed over the cortical surface ($X \in \mathbb{R}^{364 \times 1884}$). The orientation of the candidate dipole is assumed known and normal to the cortical mantle.

The measurements for $q = 1$ (single time measurements) are selected 75 ms after the stimulus onset, and between 60 and 115 ms after the stimulus for $q = 34$. This time interval corresponds to the main cortical response to the auditory stimulation.

For a number $t$ of repetitions of experiment ($t$ ranging from 2 to 56), we create an observation matrix $Y_t$ by averaging the first $t$ trials. By doing so, the noise standard deviations of each block should be proportional to $1/\sqrt{t}$. We then run the multi-task SBHCL with fixed $\lambda$, equal to 3% of $\lambda_{max}$. Figure 3 shows the noise standard deviation estimated by the multi-task SBHCL, when ran on a single time instant (single task) and 34 time instants.

We can see that the estimated values are plausible: they have the correct orders of magnitude, as well as the expected $1/\sqrt{t}$ decrease. We also see that taking more tasks into account leads to more stable noise estimation. Indeed using more tasks (here more time instants) reduces the variance of the estimation.

## 6 Conclusion

This work proposes the multi-task Smoothed Generalized Concomitant Lasso, a new sparse regression estimator designed to deal with heterogeneous observations coming from different origins and corrupted by different levels of noise. Despite the joint estimation of the regression coefficients as well as the noise level, the problem considered is jointly convex, thus guaranteeing global convergence which one can check by duality gap certificates. The efficient BCD strategy we proposed leads to a computational complexity not higher than the one observed for a classic sparse regression model, while solving a fundamental practical problem. Indeed with the SBHCL, the regularization parameter is less sensitive to the noise level of each combined modality, making it easier to tune across experimental conditions and datasets. Finally, thanks to the flexibility of our model, better prediction performance and support recovery are achieved *w.r.t.* traditional homoscedastic estimators.

## Acknowledgments

This work was funded by the ERC Starting Grant SLAB ERC-YStG-676943.

# A Proofs

## A.1 Dual of multi-task SGCL

Here we prove the result of Theorem 1. The following problems are equivalent:

$$\min_{B\in\mathbb{R}^{p\times q}, \Sigma \succeq \underline{\Sigma}} \frac{1}{2nq} \|Y - XB\|_{\Sigma^{-1}}^2 + \frac{1}{2n}\mathrm{Tr}(\Sigma) + \lambda \|B\|_{2,1}$$

$$= \min_{B\in\mathbb{R}^{p\times q}, \Sigma \succeq \underline{\Sigma}, Z\in\mathbb{R}^{n\times q}} \frac{1}{2nq} \|Z\|_{\Sigma^{-1}}^2 + \frac{1}{2n}\mathrm{Tr}(\Sigma) + \lambda \|B\|_{2,1} \text{ s.t. } Z = (Y - XB)$$

$$= \min_{B\in\mathbb{R}^{p\times q}, \Sigma \succeq \underline{\Sigma}, Z\in\mathbb{R}^{n\times q}} \max_{\Theta\in\mathbb{R}^{n\times q}} \underbrace{\frac{1}{2nq} \|Z\|_{\Sigma^{-1}}^2 + \frac{1}{2n}\mathrm{Tr}(\Sigma) + \lambda \|B\|_{2,1} + \lambda \langle \Theta, Y - XB - Z \rangle}_{\mathcal{L}(B,\Sigma,\Theta,Z)}$$

$$= \max_{\Theta\in\mathbb{R}^{n\times q}} \min_{\Sigma \succeq \underline{\Sigma}} \frac{1}{2n}\mathrm{Tr}(\Sigma) - \max_{Z\in\mathbb{R}^{n\times q}} \left\{ \langle \lambda\Theta, Z \rangle - \frac{1}{2nq}\langle Z, \Sigma^{-1}Z\rangle \right\} - \lambda \max_{B\in\mathbb{R}^{p\times q}} \left\{ \langle X^\top\Theta, B\rangle - \|B\|_{2,1} \right\} + \langle \lambda\Theta, Y \rangle$$

$$= \max_{\Theta\in\mathbb{R}^{n\times q}} \min_{\Sigma \succeq \underline{\Sigma}} \left\{ \frac{1}{2n}\mathrm{Tr}(\Sigma) - \frac{nq\lambda^2}{2}\mathrm{Tr}(\Theta^\top \Sigma \Theta) \right\} - \iota_{\mathcal{B}_{2,\infty}}(X^\top\Theta) + \langle \lambda\Theta, Y\rangle$$

$$= \max_{\Theta\in\mathbb{R}^{n\times q}} \min_{\Sigma \succeq \underline{\Sigma}} \frac{1}{2n}\mathrm{Tr}[(\mathrm{I} - n^2 q \lambda^2 \Theta\Theta^\top)\Sigma] - \iota_{\mathcal{B}_{2,\infty}}(X^\top\Theta) + \langle \lambda\Theta, Y\rangle$$

The fourth line is true because Slater's condition is met. Finally, we need to compute

$$\min_{\Sigma \succeq \underline{\Sigma}} \frac{1}{2n} \mathrm{Tr}[(\mathrm{I} - n^2 q \lambda^2 \Theta\Theta^\top)\Sigma] = \begin{cases} \frac{1}{2n} \mathrm{Tr}[(\mathrm{I} - n^2 q \lambda^2 \Theta\Theta^\top)\underline{\Sigma}], & \text{if } \mathrm{I} - n^2\lambda^2 \Theta\Theta^\top \succeq 0, \\ -\infty, & \text{otherwise.} \end{cases} \quad (A.1)$$

The condition $\mathrm{I}_n - n^2 q \lambda^2 \Theta\Theta^\top \succeq 0$ is equivalent to: $\|\Theta\Theta^\top\|_2 = \|\Theta\|_2^2 \leq \frac{1}{n^2 q \lambda^2}$.

Also, since $\underline{\Sigma} = \underline{\sigma}\, \mathrm{I}_n$, we have $\frac{1}{2n}\mathrm{Tr}[(\mathrm{I}_n - n^2 q \lambda^2 \Theta\Theta^\top)\underline{\Sigma}] = \frac{\underline{\sigma}}{2}\left(1 - nq\lambda^2 \|\Theta\|^2\right)$.

The link equation and the subdifferential inclusion come from first order optimality conditions on the Lagrangian:

- $\frac{\partial \mathcal{L}(\hat{B}, \hat{\Sigma}, \hat{\Theta}, \cdot)}{\partial Z}(\hat{Z}) = 0$ gives $\hat{\Theta} = \frac{1}{nq}\Sigma^{-1}\hat{Z}$.

- $\frac{\partial \mathcal{L}(\hat{B}, \hat{\Sigma}, \cdot, \hat{Z})}{\partial \Theta}(\hat{\Theta}) = 0$ gives $\hat{Z} = Y - X\hat{B}$.

- $0 \in \partial \mathcal{L}(\cdot, \hat{\Sigma}, \hat{\Theta}, \hat{Z})(\hat{B})$ gives $\lambda X^\top \hat{\Theta} \in \lambda \partial \|\cdot\|_{2,1}(\hat{B})$.

## A.2 Update of $\Sigma$ for multi-task SGCL

In this section we prove the result from Proposition 2. Let us solve

$$\arg\min_{\Sigma \in \mathbb{S}_{++}^n, \Sigma \succeq \underline{\Sigma}} \frac{1}{2nq} \mathrm{Tr}[(Y-XB)^\top \Sigma^{-1}(Y-XB)] + \frac{1}{2n}\mathrm{Tr}(\Sigma) \enspace. \quad (A.2)$$

*Proof.* Recall that we use $\underline{\Sigma} = \underline{\sigma}\, \mathrm{I}_n$. With $Z = (Y - XB)/\sqrt{q}$, then problem (A.2) is equivalent to solving

$$\arg\min_{\Sigma \in \mathbb{S}_{++}^n, \Sigma \succeq \underline{\Sigma}} \mathrm{Tr}\, Z^\top \Sigma^{-1} Z + \mathrm{Tr}(\Sigma) \enspace, \quad (A.3)$$

for which strong duality holds since Slater's conditions are met. The associated Lagrangian formulation is:

$$\min_{\Sigma \in \mathbb{S}^n} \max_{\Lambda \in \mathbb{S}_+^n} \underbrace{\mathrm{Tr}\, Z^\top \Sigma^{-1} Z + \mathrm{Tr}(\Sigma) + \mathrm{Tr}(\Lambda^\top(\underline{\Sigma} - \Sigma))}_{\mathcal{L}(\Sigma, \Lambda)} \enspace. \quad (A.4)$$

Recall that the gradient of $\Sigma \mapsto \operatorname{Tr} Z^\top \Sigma^{-1} Z$ is $-\Sigma^{-1} ZZ^\top \Sigma^{-1}$ (on $\mathbb{S}_{++}^n$), hence the first order optimality conditions read

$$\frac{\partial \mathcal{L}(\cdot, \hat{\Lambda})}{\partial \Sigma}(\hat{\Sigma}) = -\hat{\Sigma}^{-1} ZZ^\top \hat{\Sigma}^{-1} + I_n - \hat{\Lambda} = 0 ,$$
$$\Lambda^\top(\underline{\Sigma} - \hat{\Sigma}) = 0 ,$$
$$\hat{\Lambda} \in \mathbb{S}_+^n ,$$
$$\hat{\Sigma} \geq \underline{\Sigma} .$$

If $\|Z\|_2 > \underline{\sigma}$, let $U \operatorname{diag}(\lambda_1, \ldots, \lambda_r, 0, \ldots, 0) U^\top$ be an eigenvalue decomposition of $ZZ^\top$, with $r$ the rank of $ZZ^\top$, $\lambda_1 \geq \cdots \geq \lambda_r > 0$ and $UU^\top = I_n$.

For $i \in [r]$, let us defined $\mu_i = \sqrt{\lambda_i} \vee \underline{\sigma}$, $\Sigma = U \operatorname{diag}(\mu_1, \ldots, \mu_r, \underline{\sigma}, \ldots, \underline{\sigma}) U^\top$, and $\Lambda = I_n - \Sigma^{-1} ZZ^\top \Sigma^{-1}$. It is clear by construction that $\Sigma \geq \underline{\Sigma}$. We also have

$$\Lambda = U \operatorname{diag}(1, \ldots, 1) U^\top - U \operatorname{diag}(\frac{\lambda_1}{\mu_1^2}, \ldots, \frac{\lambda_r}{\mu_r^2}, 0, \ldots, 0) U^\top \tag{A.5}$$

$$= U \operatorname{diag}(1 - \frac{\lambda_1}{\mu_1^2}, \ldots, 1 - \frac{\lambda_1}{\mu_1^2}, 1, \ldots, 1) U^\top \tag{A.6}$$

$$\geq 0 , \tag{A.7}$$

where the later holds thanks to the $\mu_i$'s definition. Moreover,

$$\Lambda^\top(\Sigma - \underline{\Sigma}) = U \operatorname{diag}\left(\left(1 - \frac{\lambda_1}{\mu_1^2}\right)(\mu_1 - \underline{\sigma}), \ldots, \left(1 - \frac{\lambda_r}{\mu_r^2}\right)(\mu_r - \underline{\sigma}), 0, \ldots, 0\right) U^\top = 0$$

since $\forall i \in [r]$, $\left(1 - \frac{\lambda_i}{\mu_i^2}\right)(\mu_i - \underline{\sigma}) = 0$ (either the left term or the right term is 0 by construction).

This shows that the pair $(\Sigma, \Lambda)$ satisfies all the first order conditions on the Lagrangian, hence that $\Sigma$ is solution of Problem (A.2).

$\square$

**Remark 5.** If $\|Z\|_2 \leq \underline{\sigma}$, the pair $(\Sigma, \Lambda)$ simplifies to: $(\underline{\Sigma}, I_n - ZZ^\top / \underline{\sigma}^2)$.

## A.3 BCD update for the multi-task Smoothed Generalized Concomitant Lasso

The minimization of the multi-task Smoothed Generalized Concomitant Lasso relies on alternate minimization. To minimize the objective function *w.r.t.* B, we use cyclic block-coordinate descent. To minimize $f(B) = \frac{1}{2nq} \|Y - XB\|_{\Sigma^{-1}}^2 + \lambda \|B\|_{2,1}$ with respect to $B_j$ (the $j^{th}$ line of B), a simple majorization-minimization approach is to perform

$$B_j \leftarrow \operatorname{BST}\left(B_j - \frac{\nabla_j f(B)}{\nabla_{j,j}^2 f(B)}, \frac{\lambda}{\nabla_{j,j}^2 f(B)}\right) ,$$

where $\nabla_j f(B) \in \mathbb{R}^q$ is the partial derivative of $f$ *w.r.t.* $B_j$ and $\nabla_{j,j}^2 f(B)$ is the $j^{th}$ diagonal element of the Hessian of $f$ at B.

We have $\nabla_j f(B) = -X_j^\top \Sigma^{-1}(Y - XB)/(nq)$ and $\nabla_{j,j}^2 f(B) = \|X_j\|_{\Sigma^{-1}}^2/(nq)$, leading to:

$$B_j \leftarrow \operatorname{BST}\left(B_j + \frac{X_j^\top \Sigma^{-1}(Y - XB)}{\|X_j\|_{\Sigma^{-1}}^2}, \frac{\lambda nq}{\|X_j\|_{\Sigma^{-1}}^2}\right) .$$

## A.4 Dual of multi-task SBHCL

Here we prove the result of Theorem 2. The following problems are equivalent:

$$\min_{B\in\mathbb{R}^{p\times q}, \sigma_1 \geq \underline{\sigma}_1,\ldots,\sigma_K \geq \underline{\sigma}_K} \frac{1}{n} \sum_{k=1}^{K} \left( \frac{\|Y^k - X^k B\|^2}{2q\sigma_k} + \frac{n_k \sigma_k}{2} \right) + \lambda \|B\|_{2,1}$$

$$= \min_{B\in\mathbb{R}^p, \sigma_1 \geq \underline{\sigma}_1,\ldots,\sigma_K \geq \underline{\sigma}_K} \frac{1}{n} \sum_{k=1}^{K} \left( \frac{\|Z^k\|^2}{2q\sigma_k} + \frac{n_k \sigma_k}{2} \right) + \lambda \|B\|_{2,1} \text{ s.t. } Z = Y - XB$$

$$= \min_{B\in\mathbb{R}^p, \sigma_1 \geq \underline{\sigma}_1,\ldots,\sigma_K \geq \underline{\sigma}_K} \max_{\Theta \in \mathbb{R}^{n\times q}} \underbrace{\frac{1}{n} \sum_{k=1}^{K} \left( \frac{\|Z^k\|^2}{2q\sigma_k} + \frac{n_k \sigma_k}{2} \right) + \lambda \|B\|_{2,1} + \langle \lambda \Theta, Y - XB - Z \rangle}_{\mathcal{L}(B, \sigma_1,\ldots,\sigma_K, \Theta, Z)}$$

$$= \max_{\Theta \in \mathbb{R}^{n\times q}} \min_{\sigma_1 \geq \underline{\sigma}_1,\ldots,\sigma_K \geq \underline{\sigma}_K} \frac{\sum_{k=1}^{K} n_k \sigma_k}{2n} - \max_{Z\in\mathbb{R}^{n\times q}} \left\{ \langle \lambda \Theta, Z \rangle - \sum_{k=1}^{K} \frac{\|Z^k\|^2}{2nq\sigma_k} \right\} - \lambda \max_{B\in\mathbb{R}^p} \left\{ \langle X^\top \Theta, B \rangle - \|B\|_{2,1} \right\} + \langle \lambda \Theta, Y \rangle$$

$$= \max_{\Theta \in \mathbb{R}^{n\times q}} \min_{\sigma_1 \geq \underline{\sigma}_1,\ldots,\sigma_K \geq \underline{\sigma}_K} \frac{\sum_{k=1}^{K} n_k \sigma_k}{2n} - \frac{nq\lambda^2}{2} \sum_{k=1}^{K} \sigma_k \|\Theta^k\|^2 - \iota_{\mathcal{B}_{2,\infty}}(X^\top \Theta) + \langle \lambda \Theta, Y \rangle$$

$$= \max_{\Theta \in \mathbb{R}^{n\times q}} \min_{\sigma_1 \geq \underline{\sigma}_1,\ldots,\sigma_K \geq \underline{\sigma}_K} \frac{1}{2n} \sum_{k=1}^{K} \sigma_k \left( n_k - n^2 q \lambda^2 \|\Theta^k\|^2 \right) - \iota_{\mathcal{B}_{2,\infty}}(X^\top \Theta) + \langle \lambda \Theta, Y \rangle.$$

The fourth line is true because Slater's condition is met, hence we can permute min and max thanks to strong duality. Finally we obtain the targeted dual problem formulation since

$$\min_{\sigma_1 \geq \underline{\sigma}_1,\ldots,\sigma_K \geq \underline{\sigma}_K} \frac{1}{2n} \sum_{k=1}^{K} \sigma_k \left( n_k - n^2 q \lambda^2 \|\Theta^k\|^2 \right) =$$

$$\begin{cases} \frac{1}{2n} \sum_{k=1}^{K} \underline{\sigma}_k \left( n_k - n^2 q \lambda^2 \|\Theta^k\|^2 \right), & \text{if } \forall k \in [K], n_k - n^2 q \lambda^2 \|\Theta^k\|^2 \geq 0, \\ -\infty, & \text{otherwise.} \end{cases}$$

The condition $\forall k \in [K], n_k - n^2 q \lambda^2 \|\Theta^k\|^2 \geq 0$ is equivalent to $\forall k \in [K], \|\Theta^k\| \leq \frac{\sqrt{n_k}}{n\lambda\sqrt{q}}$.
Finally, the primal-dual link equation follows directly from Fermat's rule:

$$\frac{\partial \mathcal{L}(\hat{B}, \hat{\sigma}_1,\ldots,\hat{\sigma}_K, \cdot, \hat{Z})}{\partial \Theta}(\hat{\Theta}) = Y - X\hat{B} - \hat{Z} = 0,$$

$$\forall k \in [K], \quad \frac{\partial \mathcal{L}(\hat{B},\ldots,\hat{\sigma}_k,\ldots,\hat{\Theta}, \cdot)}{\partial Z^k}(\hat{Z}) = \frac{\hat{Z}^k}{n\hat{\sigma}_k} - \lambda \hat{\Theta}^k = 0,$$

so the dual solution $\hat{\Theta}$ satisfies: $\forall k \in [K], \hat{\Theta}^k = \frac{1}{n\lambda\hat{\sigma}_k}(Y - X^k \hat{B})$.

# B  Single task case ($q = 1$)

Since this work might be of interest to (single task) Lasso users, and because the various formulas are simpler when $q = 1$, we include this case here.

## B.1 Dual formulation of the general problem

$$\min_{\beta \in \mathbb{R}^p, \Sigma \geq \underline{\Sigma}} \frac{1}{2n}(y - X\beta)^\top \Sigma^{-1}(y - X\beta) + \frac{1}{2n}\text{Tr}(\Sigma) + \lambda \|\beta\|_1$$

$$= \min_{\beta \in \mathbb{R}^p, z \in \mathbb{R}^n, \Sigma \geq \underline{\Sigma}} \frac{1}{2n} z^\top \Sigma^{-1} z + \frac{1}{2n}\text{Tr}(\Sigma) + \lambda \|\beta\|_1 \text{ s.t. } z = y - X\beta$$

$$= \min_{\beta \in \mathbb{R}^p, z \in \mathbb{R}^n, \Sigma \geq \underline{\Sigma}} \max_{\theta \in \mathbb{R}^n} \underbrace{\frac{1}{2n} z^\top \Sigma^{-1} z + \frac{1}{2n}\text{Tr}(\Sigma) + \lambda \|\beta\|_1 + \langle \lambda\theta, y - X\beta - z \rangle}_{\mathcal{L}(\beta, \Sigma, \theta, z)}$$

$$= \max_{\theta \in \mathbb{R}^n} \min_{\Sigma \geq \underline{\Sigma}} \frac{\text{Tr}(\Sigma)}{2n} - \max_{z \in \mathbb{R}^n}\left\{\langle \lambda\theta, z\rangle - \frac{1}{2n} z^\top \Sigma^{-1} z\right\} - \lambda \max_{\beta \in \mathbb{R}^p}\left\{\langle X^\top \theta, \beta\rangle - \|\beta\|_1\right\} + \langle \lambda\theta, y\rangle$$

$$= \max_{\theta \in \mathbb{R}^n} \min_{\Sigma \geq \underline{\Sigma}} \left\{\frac{1}{2n}\text{Tr}(\Sigma) - \frac{n\lambda^2}{2}\theta^\top \Sigma \theta\right\} - \iota_{\mathcal{B}_\infty}(X^\top \theta) + \langle \lambda\theta, y\rangle$$

$$= \max_{\theta \in \mathbb{R}^n} \min_{\Sigma \geq \underline{\Sigma}} \frac{\text{Tr}[(I_n - n^2\lambda^2\theta\theta^\top)^\top \Sigma]}{2n} - \iota_{\mathcal{B}_\infty}(X^\top \theta) + \langle \lambda\theta, y\rangle.$$

The fourth line is true because Slater's condition is met, hence we can permute min and max thanks to strong duality. Finally we obtain the dual problem since

$$\min_{\Sigma \geq \underline{\Sigma}} \frac{\text{Tr}[(I_n - n^2\lambda^2\theta\theta^\top)^\top \Sigma]}{2n} = \begin{cases} \frac{\text{Tr}[(I_n - n^2\lambda^2\theta\theta^\top)^\top \underline{\Sigma}]}{2n}, & \text{if } I_n - n^2\lambda^2\theta\theta^\top \geq 0, \\ -\infty, & \text{otherwise.} \end{cases}$$

Note that since the matrix $n^2\lambda^2\theta\theta^\top$ is rank one, the condition $I_n - n^2\lambda^2\theta\theta^\top \geq 0$ is equivalent to the simpler: $1 \geq n^2\lambda^2 \|\theta\|^2$, i.e., $1 \geq n\lambda \|\theta\|$

For Fermat's condition, let us use the same Lagragian notation as above, and denote

$$\left(\hat{\beta}, \hat{\Sigma}, \hat{\theta}, \hat{z}\right) \in \arg\min_{\beta \in \mathbb{R}^p, z \in \mathbb{R}^n, \Sigma \geq \underline{\Sigma}} \max_{\theta \in \mathbb{R}^n} \mathcal{L}(\beta, \Sigma, \theta, z).$$

The primal-dual link equation follows directly from Fermat's rule:

$$\frac{\partial \mathcal{L}(\hat{\beta}, \hat{\Sigma}, \cdot, \hat{z})}{\partial \theta}(\hat{\theta}) = y - X\hat{\beta} - \hat{z} = 0,$$

$$\frac{\partial \mathcal{L}(\hat{\beta}, \hat{\Sigma}, \hat{\theta}, \cdot)}{\partial z}(\hat{z}) = \frac{\hat{\Sigma}^{-1}\hat{z}}{n} - \lambda\hat{\theta} = 0.$$

## B.2 Dual formulation for the block homoscedastic setting

$$\min_{\beta \in \mathbb{R}^p, \sigma_1 \geq \underline{\sigma}_1, \ldots, \sigma_K \geq \underline{\sigma}_K} \frac{1}{n} \sum_{k=1}^{K} \left( \frac{\|y^k - X^k\beta\|^2}{2\sigma_k} + \frac{n_k \sigma_k}{2} \right) + \lambda \|\beta\|_1$$

$$\min_{\beta \in \mathbb{R}^p, \sigma_1 \geq \underline{\sigma}_1, \ldots, \sigma_K \geq \underline{\sigma}_K} \frac{1}{n} \sum_{k=1}^{K} \left( \frac{\|z^k\|^2}{2\sigma_k} + \frac{n_k \sigma_k}{2} \right) + \lambda \|\beta\|_1 \quad \text{s.t. } z = y - X\beta$$

$$\min_{\beta \in \mathbb{R}^p, \sigma_1 \geq \underline{\sigma}_1, \ldots, \sigma_K \geq \underline{\sigma}_K} \max_{\theta \in \mathbb{R}^n} \underbrace{\frac{1}{n} \sum_{k=1}^{K} \left( \frac{\|z^k\|^2}{2\sigma_k} + \frac{n_k \sigma_k}{2} \right) + \lambda \|\beta\|_1 + \langle \lambda\theta, y - X\beta - z \rangle}_{\mathcal{L}(\beta, \sigma_1, \ldots, \sigma_K, \theta, z)}$$

$$= \max_{\theta \in \mathbb{R}^n} \min_{\sigma_1 \geq \underline{\sigma}_1, \ldots, \sigma_K \geq \underline{\sigma}_K} \frac{\sum_{k=1}^{K} n_k \sigma_k}{2n} - \max_{z \in \mathbb{R}^n} \left\{ \langle \lambda\theta, z \rangle - \sum_{k=1}^{K} \frac{1}{2n\sigma_k} \|z^k\|^2 \right\} - \lambda \max_{\beta \in \mathbb{R}^p} \left\{ \langle X^\top \theta, \beta \rangle - \|\beta\|_1 \right\} + \langle \lambda\theta, y \rangle$$

$$= \max_{\theta \in \mathbb{R}^n} \min_{\sigma_1 \geq \underline{\sigma}_1, \ldots, \sigma_K \geq \underline{\sigma}_K} \frac{\sum_{k=1}^{K} n_k \sigma_k}{2n} - \frac{n\lambda^2}{2} \sum_{k=1}^{K} \sigma_k \|\theta^k\|^2 - \iota_{\mathcal{B}_\infty}(X^\top \theta) + \langle \lambda\theta, y \rangle$$

$$= \max_{\theta \in \mathbb{R}^n} \min_{\sigma_1 \geq \underline{\sigma}_1, \ldots, \sigma_K \geq \underline{\sigma}_K} \frac{1}{2n} \sum_{k=1}^{K} \sigma_k \left( n_k - n^2 \lambda^2 \|\theta^k\|^2 \right) - \iota_{\mathcal{B}_\infty}(X^\top \theta) + \langle \lambda\theta, y \rangle.$$

The fourth line is true because Slater's condition is met, hence we can permute min and max thanks to strong duality. Finally we obtain the dual problem since

$$\min_{\sigma_1 \geq \underline{\sigma}_1, \ldots, \sigma_K \geq \underline{\sigma}_K} \frac{1}{2n} \sum_{k=1}^{K} \sigma_k \left( n_k - n^2 \lambda^2 \|\theta^k\|^2 \right) =$$

$$\begin{cases} \frac{1}{2n} \sum_{k=1}^{K} \underline{\sigma}_k \left( n_k - n^2 \lambda^2 \|\theta^k\|^2 \right), & \text{if } \forall k \in [K], n_k - n^2 \lambda^2 \|\theta^k\|^2 > 0, \\ -\infty, & \text{otherwise.} \end{cases}$$

This is equivalent to $\forall k \in [K], n_k \geq n^2 \lambda^2 \|\theta\|^2$.

For the Fermat condition, let us use the same Lagragian notation as above, and denote

$$\left( \hat{\beta}, \hat{\sigma}_1, \ldots, \hat{\sigma}_K, \hat{\theta}, \hat{z} \right) \in \operatorname*{arg\,min}_{\beta \in \mathbb{R}^p, z \in \mathbb{R}^n, \sigma_1 \geq \underline{\sigma}_1, \cdots, \sigma_K \geq \underline{\sigma}_K} \max_{\theta \in \mathbb{R}^n} \mathcal{L}(\beta, \sigma_1, \ldots, \sigma_K, \theta, z).$$

The primal-dual link equation follows directly from the Fermat's rule:

$$\frac{\partial \mathcal{L}(\hat{\beta}, \hat{\sigma}_1, \ldots, \hat{\sigma}_K, \cdot, \hat{z})}{\partial \theta}(\hat{\theta}) = y - X\hat{\beta} - \hat{z} = 0,$$

$$\forall k \in [K], \quad \frac{\partial \mathcal{L}(\hat{\beta}, \ldots, \hat{\sigma}_k, \ldots, \hat{\theta}, \cdot)}{\partial z^k}(\hat{z}) = \frac{\hat{z}^k}{n\hat{\sigma}_k} - \lambda \hat{\theta}^k = 0.$$

## B.3 Square root of covariance update in the single task case

In this part, we simplify the results from Proposition 2 when $q = 1$, more precisely we show that the matrix

$$\underline{\Sigma} + \frac{(\|y - X\beta\| - \underline{\sigma})_+}{\|y - X\beta\|^2}(y - X\beta)(y - X\beta)^\top \tag{B.1}$$

is a solution of the problem:

$$\underset{\Sigma \in \mathbb{S}_{++}^n, \Sigma \geq \underline{\Sigma}}{\arg\min} \frac{1}{2n}(y - X\beta)^\top \Sigma^{-1}(y - X\beta) + \frac{1}{2n}\text{Tr}(\Sigma) \ . \tag{B.2}$$

*Proof.* Recall that we use $\underline{\Sigma} = \underline{\sigma} \, I_n$. With $z = y - X\beta$, then problem (B.2) is equivalent to solving

$$\underset{\Sigma \in \mathbb{S}_{++}^n, \Sigma \geq \underline{\Sigma}}{\arg\min} \ z^\top \Sigma^{-1} z + \text{Tr}(\Sigma) \ . \tag{B.3}$$

Let us recall the Lagrangian formulation associated:

$$\min_{\Sigma \in \mathbb{S}^n} \max_{\Lambda \in \mathbb{S}_+^n} \underbrace{z^\top \Sigma^{-1} z + \text{Tr}(\Sigma) + \text{Tr}(\Lambda^\top(\underline{\Sigma} - \Sigma))}_{\mathcal{L}(\Sigma, \Lambda)} \ . \tag{B.4}$$

Recall that the gradient of $\Sigma \mapsto z^\top \Sigma^{-1} z$ is $-\Sigma^{-1} z z^\top \Sigma^{-1}$ hence the first order optimality conditions read

$$\frac{\partial \mathcal{L}(\cdot, \hat{\Lambda})}{\partial \Sigma}(\hat{\Sigma}) = -\hat{\Sigma}^{-1} z z^\top \hat{\Sigma}^{-1} + I_n - \hat{\Lambda} = 0 \ ,$$
$$\Lambda^\top(\underline{\Sigma} - \hat{\Sigma}) = 0 \ ,$$
$$\hat{\Lambda} \in \mathbb{S}_+^n \ ,$$
$$\hat{\Sigma} \geq \underline{\Sigma} \ .$$

- If $\|z\| \leq \underline{\sigma}$, it is trivial to check that the pair $(\underline{\Sigma}, I_n - zz^\top / \underline{\sigma}^2)$ satisfies these conditions.

- If $\|z\| > \underline{\sigma}$, let us check that for a well-chosen $\gamma$, the matrix $\Sigma = \underline{\Sigma} + \gamma z z^\top$ satisfies the first equality. Recall the Sherman-Morrison formula:

$$(\underline{\sigma} \, I_n + \gamma z z^\top)^{-1} = \underline{\Sigma}^{-1} - \gamma \frac{zz^\top / \underline{\sigma}^2}{1 + \gamma z^\top z / \underline{\sigma}} \ . \tag{B.5}$$

$$\Lambda = I_n - \Sigma^{-1} z z^\top \Sigma^{-1}$$
$$= I_n - \left( \underline{\Sigma}^{-1} - \gamma \frac{zz^\top / \underline{\sigma}^2}{1 + \gamma z^\top z / \underline{\sigma}} \right) zz^\top \left( \underline{\Sigma}^{-1} - \gamma \frac{zz^\top / \underline{\sigma}^2}{1 + \gamma z^\top z \underline{\sigma}} \right)$$
$$= I_n - \left( zz^\top / \underline{\sigma}^2 - 2\gamma \frac{zz^\top zz^\top / \underline{\sigma}^3}{1 + \gamma z^\top z / \underline{\sigma}} + \gamma^2 \frac{zz^\top zz^\top zz^\top / \underline{\sigma}^4}{(1 + \gamma z^\top z / \underline{\sigma})^2} \right)$$
$$= I_n - \left( 1 - 2\gamma \frac{z^\top z / \underline{\sigma}}{1 + \gamma z^\top z / \underline{\sigma}} + \gamma^2 \left( \frac{z^\top z / \underline{\sigma}}{1 + \gamma z^\top z / \underline{\sigma}} \right)^2 \right) zz^\top / \underline{\sigma}^2$$
$$= I_n - \left( 1 - \gamma \frac{z^\top z / \underline{\sigma}}{1 + \gamma z^\top z / \underline{\sigma}} \right)^2 zz^\top / \underline{\sigma}^2 \ .$$

In particular, provided

$$\left( 1 - \gamma \frac{z^\top z / \underline{\sigma}}{1 + \gamma z^\top z / \underline{\sigma}} \right)^2 \|z\|^2 \leq \underline{\sigma}^2 \ , \tag{B.6}$$

that is,

$$\gamma \geq \frac{\|z\| - \underline{\sigma}}{\|z\|^2} \tag{B.7}$$

then one can check that $\Lambda \succeq 0$. Moreover, as $\Sigma = \underline{\Sigma} + \gamma zz^\top$,

$$\mathrm{Tr}(\Lambda^\top(\underline{\Sigma} - \hat{\Sigma})) = -\gamma \, \mathrm{Tr}\left(zz^\top - \left(1 - \gamma \frac{z^\top z/\underline{\sigma}}{1 + \gamma z^\top z/\underline{\sigma}}\right)^2 zz^\top zz^\top/\underline{\sigma}^2\right) \tag{B.8}$$

$$= -\gamma \|z\|^2 + \gamma \left(1 - \gamma \frac{z^\top z/\underline{\sigma}}{1 + \gamma z^\top z/\underline{\sigma}}\right)^2 \|z\|^4/\underline{\sigma}^2 \tag{B.9}$$

It is easy to check that if we choose $\gamma = \dfrac{\|z\| - \underline{\sigma}}{\|z\|^2}$, then $\mathrm{Tr}(\Lambda^\top(\underline{\Sigma} - \hat{\Sigma})) = 0$, and so $\underline{\Sigma} + \dfrac{\|z\| - \underline{\sigma}}{\|z\|^2} zz^\top$ is a solution of the problem defined in (B.2).

A general formulation encompassing both cases is thus summarized by $\hat{\Sigma} = \underline{\Sigma} + \dfrac{(\|z\| - \underline{\sigma})_+}{\|z\|^2} zz^\top$.

$\square$

### B.4 BCD update for the SGCL

In the context where $q = 1$, the BCD method reduces in our context to a Coordinate Descent (CD) approach. At each CD update, a coordinate $j \in [p]$ is picked and the objective function is minimized with respect to $\beta_j$, all other variables remaining fixed. Again we remind that the CD update to minimize a "smooth convex + $\ell_1$" function $\beta \mapsto f(\beta) + \lambda\|\beta\|_1$ is

$$\beta_j \leftarrow \mathrm{ST}\left(\beta_j - \frac{\nabla_j f(\beta)}{\nabla^2_{j,j} f(\beta)}, \frac{\lambda}{\nabla^2_{j,j} f(\beta)}\right), \tag{B.10}$$

where $\nabla_j f(\beta)$ is the partial derivative of $f$ at $\beta$, and $\nabla^2_{j,j} f(\beta)$ is the $j^{th}$ diagonal element of the Hessian of $f$ at $\beta$.

With the choice $f(\beta) = \frac{1}{2n}(y - X\beta)^\top \Sigma^{-1}(y - X\beta)$ on can check that $\nabla_j f(\beta) = -X_j^\top \Sigma^{-1}(y - X\beta)/n$ and $\nabla^2_{j,j} f(\beta) = X_j^\top \Sigma^{-1} X_j/n$, leading to the following update

$$\beta_j \leftarrow \mathrm{ST}\left(\beta_j + \frac{X_j^\top \Sigma^{-1}(y - X\beta)}{X_j^\top \Sigma^{-1} X_j}, \frac{\lambda n}{X_j^\top \Sigma^{-1} X_j}\right). \tag{B.11}$$

Writing, $r_{\mathrm{int}} = y - \sum_{m \neq j} X_m \beta_m$ this can also be rewritten:

$$\beta_j \leftarrow \mathrm{ST}\left(\frac{X_j^\top \Sigma^{-1} r_{\mathrm{int}}}{X_j^\top \Sigma^{-1} X_j}, \frac{\lambda n}{X_j^\top \Sigma^{-1} X_j}\right). \tag{B.12}$$

**Algorithm 3** ALTERNATE MIN. FOR SINGLE TASK MULTI-TASK SBHCL

**input :** $X, y, \underline{\Sigma}^{-1}, \lambda$
**init  :** $\beta = 0_p, \Sigma^{-1} = \underline{\Sigma}^{-1}$
**for** it $= 1, \ldots, T$ **do**
$\quad$ **if** it $\equiv 1 \pmod{f}$ **then**
$\quad\quad \gamma = \frac{(\|r\|/\underline{\sigma} - 1)_+}{\|r\|^2/\underline{\sigma}}$
$\quad\quad \Sigma^{-1} \leftarrow \underline{\Sigma}^{-1} - \gamma \frac{rr^\top/\underline{\sigma}^2}{1 + \gamma r^\top r/\underline{\sigma}}$ $\quad$ // noise matrix update
$\quad\quad$ **for** $j = 1, \ldots, p$ **do**
$\quad\quad\quad L_j = X_j^\top \Sigma^{-1} X_j$
$\quad$ **for** $j = 1, \ldots, p$ **do**
$\quad\quad r \leftarrow r + X_j \beta_j$ $\quad$ // partial residual update
$\quad\quad \beta_j \leftarrow \text{ST}\left(\frac{X_j^\top \Sigma^{-1} r}{L_j}, \frac{\lambda n}{L_j}\right)$ $\quad$ // soft-thresholding
$\quad\quad r \leftarrow r - X_j \beta_j$ $\quad$ // residual update
**return** $\beta, \Sigma$

### B.5 CD update rule for the Block homoscedastic model

At each coordinate descent step, a coordinate $j \in [p]$ is picked and the objective function is minimized with respect to $\beta_j$, all other variables remaining fixed. The new value $\beta_j$ satisfies

$$\begin{aligned}
\hat{\beta}_j &= \arg\min_{\beta_j \in \mathbb{R}} \frac{1}{2n} \sum_{k=1}^K \frac{\|y^k - X^k \beta\|^2}{\sigma_k} + \lambda |\beta_j|. \\
&= \arg\min_{\beta_j \in \mathbb{R}} \frac{1}{2n} \sum_{k=1}^K \frac{\|y^k - \sum_{m \neq j} X_m^k \beta_m - X_j^k \beta_j\|^2}{\sigma_k} + \lambda |\beta_j|. \\
&= \arg\min_{\beta_j \in \mathbb{R}} \frac{1}{2}\left(\beta_j^2 \sum_{k=1}^K \frac{L_{k,j}}{\sigma_k} - 2\beta_j \sum_{k=1}^K \frac{X_j^{k\top} r_{\text{int}}^k}{\sigma_k}\right) + \lambda n |\beta_j| \\
&= \arg\min_{\beta_j \in \mathbb{R}} \frac{1}{2}\left(\beta_j^2 \sum_{k=1}^K \frac{L_{k,j}}{\sigma_k} - 2\beta_j \sum_{k=1}^K \frac{X_j^{k\top} r_{\text{int}}^k}{\sigma_k}\right) + \lambda n |\beta_j| \\
&= \arg\min_{\beta_j \in \mathbb{R}} \frac{1}{2}\left(\beta_j^2 - 2\beta_j \frac{\sum_{k=1}^K \frac{X_j^{k\top} r_{\text{int}}^k}{\sigma_k}}{\sum_{k=1}^K \frac{L_{k,j}}{\sigma_k}}\right) + \frac{\lambda n}{\sum_{k=1}^K \frac{L_{k,j}}{\sigma_k}} |\beta_j| \\
&= \text{ST}\left(\sum_{k=1}^K \frac{X_j^{k\top} r_{\text{int}}^k}{\sigma_k}, \lambda n\right) \Big/ \sum_{k=1}^K \frac{L_{k,j}}{\sigma_k}.
\end{aligned}$$

where we use $r_{\text{int}}^k = y^k - \sum_{m \neq j} X_m^k \beta_m$.

### B.6 Algorithm

The simplified version of Algorithm 1 when $q = 1$ is provided in Algorithm 3.

## C Additional experiments

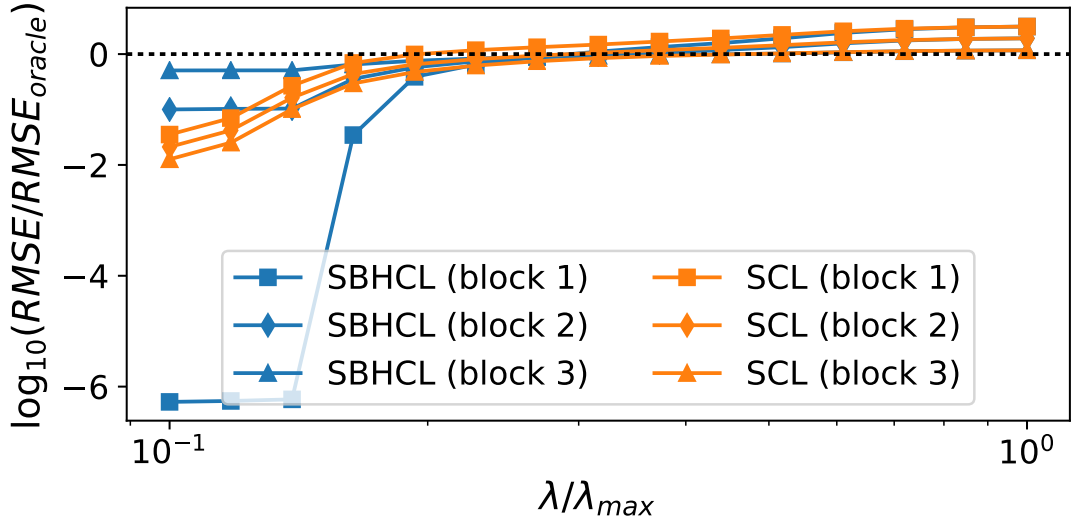
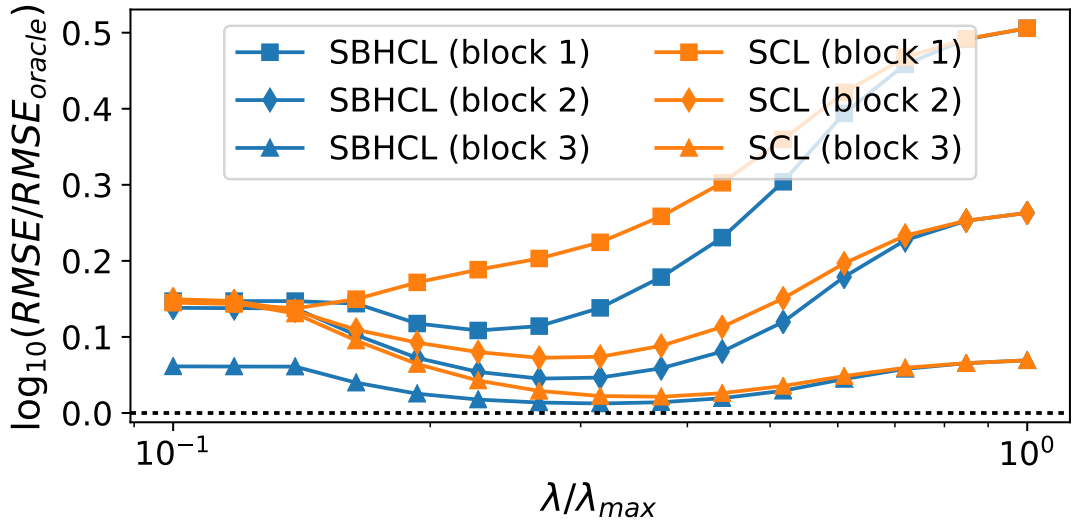

Figure 1: RMSE normalized by oracle RMSE, per block, for the multi-task SBHCL and Smooth Concomitant Lasso (SCL), on training (top) and testing (bottom) set, for various values of $\lambda$. The greatest flexibility of the block homoscedastic model enables the multi-task Smoothed Block Homoscedastic Concomitant Lasso to reach a lower RMSE on every block of the test set.

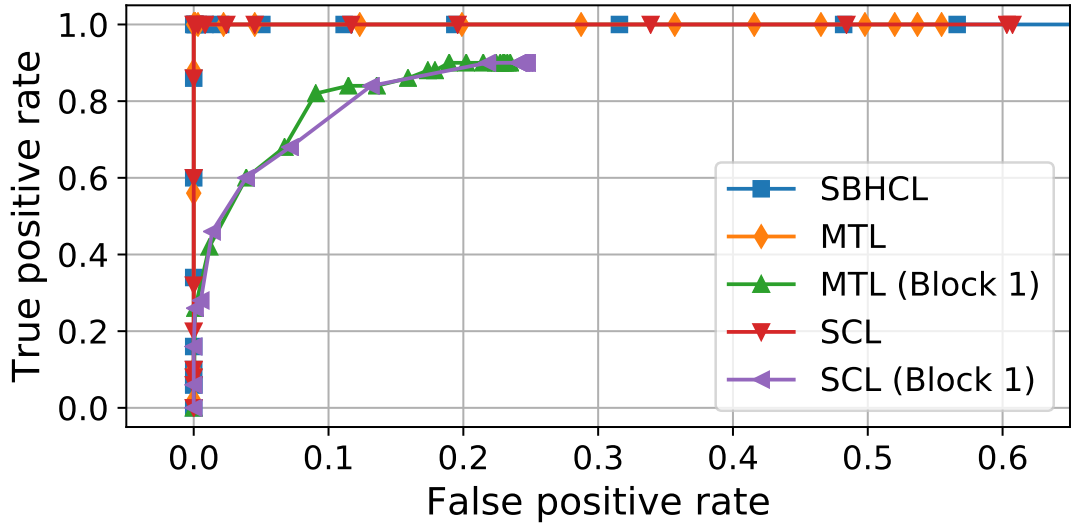
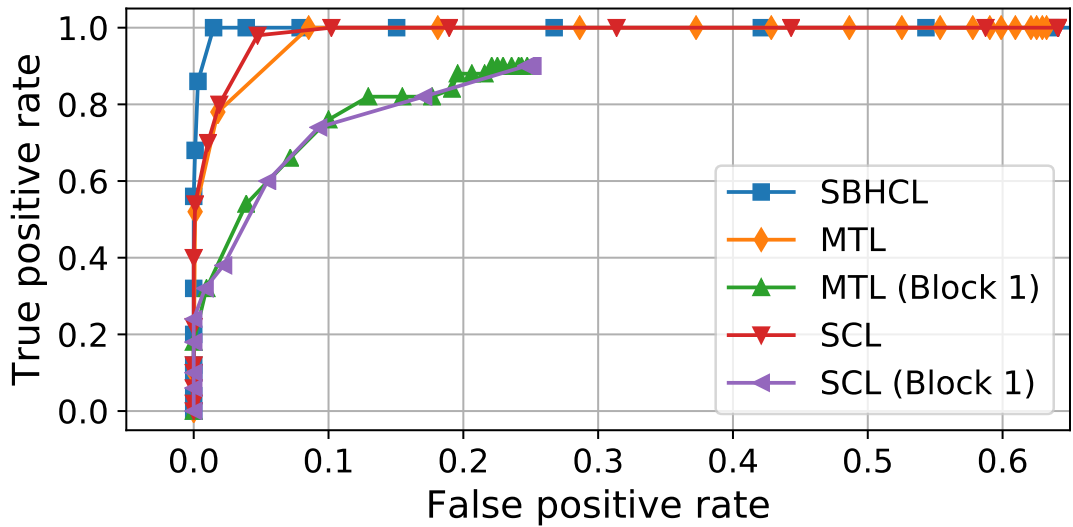
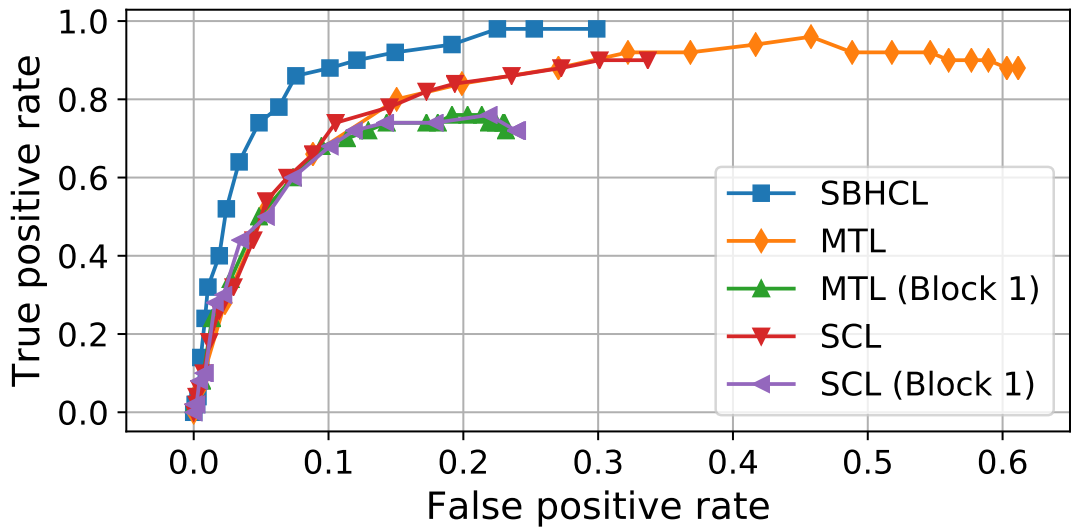

Figure 2: ROC curves of true support recovery for the SBHCL, the MTL and SCL on all blocks, and the MTL and SCL on the least noisy block. Top: SNR = 5, $\rho = 0.1$, middle: SNR=1, $\rho = 0.1$, bottom: SNR = 1, $\rho = 0.9$.

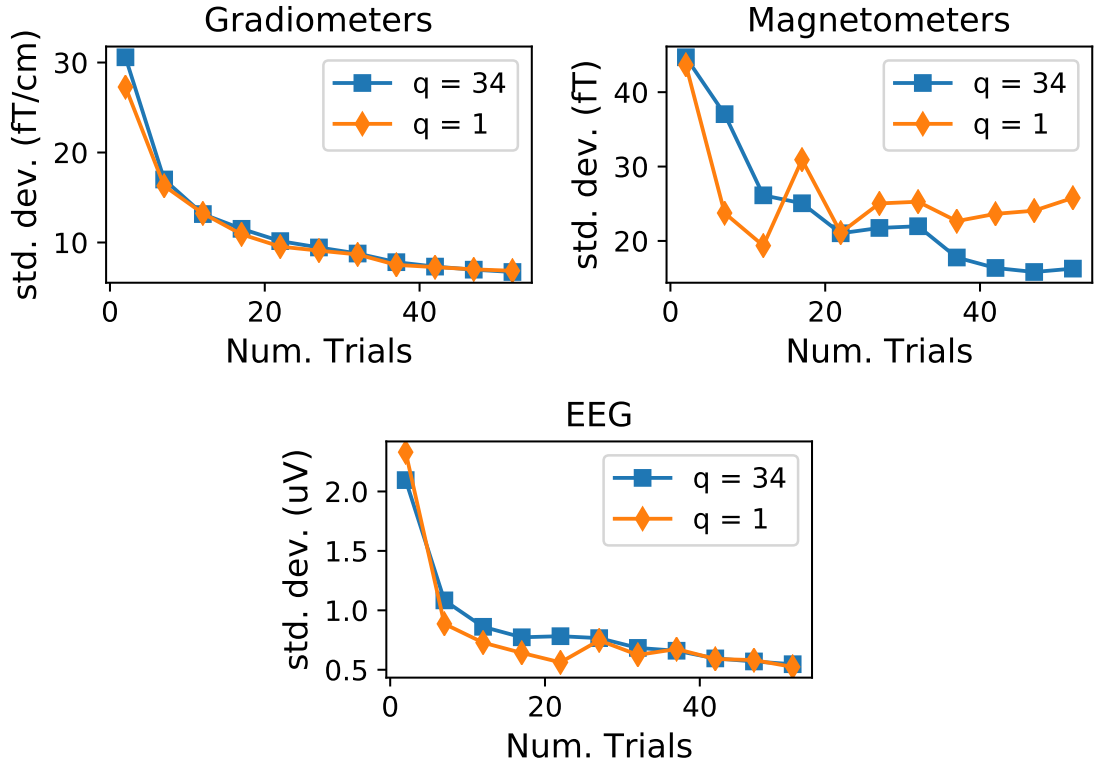

Figure 3: Noise standard deviation estimated on auditory data for $q = 1$ and $q = 34$ time instants using the SBHCL estimator. Data consist of combined MEG gradiometers ($n_1 = 203$ sensors) and magnetometers ($n_2 = 102$ sensors), as well as EEG ($n_3 = 59$ sensors). We used $\lambda = 0.03\lambda_{max}$.

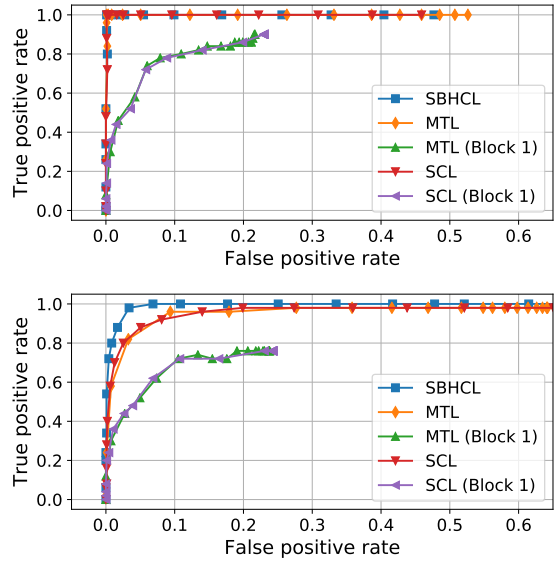

Figure 4: ROC curves of true support recovery for the SBHCL, the MTL and SCL on all blocks, and the MTL and SCL on the least noisy block. Top: SNR=5, $\rho = 0.7$, bottom: SNR=1, $\rho = 0.7$.

23